\documentclass[lettersize,journal]{IEEEtran}
\usepackage{amsmath,amsfonts,amssymb}
\usepackage{algorithmic}
\usepackage{algorithm}
\usepackage{array}
\usepackage[caption=false,font=normalsize,labelfont=sf,textfont=sf]{subfig}
\usepackage{textcomp,booktabs,multirow,multicol}
\usepackage{stfloats}
\usepackage{url}
\usepackage{verbatim}
\usepackage{graphicx}
\usepackage{cite}
\begin{document}

\title{Hessian-Free Second-Order\\ Adversarial Examples for Adversarial Learning}

\author{Yaguan Qian*, Yuqi Wang, Bin Wang, Zhaoquan Gu, Yuhan Guo and Wassim Swaileh
\thanks{Yaguan Qian, Yuqi Wang and Yuhan Guo are with the School of Big Data Science, Zhejiang University of Science and Technology, Hangzhou, China.}
\thanks{Bin Wang is with Zhejiang Key Laboratory of Multi-dimensional Perception Technology, Application, and Cybersecurity, Hikvision Digital Technology Co., Ltd, Hangzhou, China.}
\thanks{Zhaoquan Gu is with the Cyberspace Institute of Advanced Technology, Guangzhou University, Guangzhou, China.}
\thanks{Wassim Swaileh is with the ETIS Research Laboratory,CY Cergy Paris University, Paris, France.}
}

\markboth{Manuscript submitted to IEEE Transactions on Image Processing}%
{Qian \MakeLowercase{\textit{et al.}}: Hessian-Free Second-Order Adversarial Examples for Adversarial Learning}


\maketitle

\begin{abstract}
Recent studies show deep neural networks (DNNs) are extremely vulnerable to the elaborately designed adversarial examples. Adversarial learning with those adversarial examples has been proved as one of the most effective methods to defend against such an attack. At present, most existing adversarial examples generation methods are based on first-order gradients, which can hardly further improve models' robustness, especially when facing second-order adversarial attacks. Compared with first-order gradients, second-order gradients provide a more accurate approximation of the loss landscape with respect to natural examples. Inspired by this, our work crafts second-order adversarial examples and uses them to train DNNs. Nevertheless, second-order optimization involves time-consuming calculation for Hessian-inverse. We propose an approximation method through transforming the problem into an optimization in the Krylov subspace, which remarkably reduce the computational complexity to speed up the training procedure. Extensive experiments conducted on the MINIST and CIFAR-10 datasets show that our adversarial learning with second-order adversarial examples outperforms other fisrt-order methods, which can improve the model robustness against a wide range of attacks.
\end{abstract}


\section{Inroduction}
\label{sec:intro}
\IEEEPARstart{C}{onvolutional} neural networks (CNNs) have been successfully applied in many image-related tasks, such as image classification \cite{image_class02,image_class01}, target detection \cite{obj_dec}, and super-resolution \cite{super_res}. However, CNNs' vulnerability to adversarial examples has drawn great attention in the computer vision community \cite{property_adv,FGSM}. In general, an \textit{adversarial example} is an image added by an imperceptible perturbation, which can successfully fool a classifier. The existence of adversarial examples will lead to disastrous consequences, especially in safety-sensitive scenarios, such as industrial process control \cite{industry_control}, face recognition \cite{face}, and automatic drive systems \cite{auto_drive}. Numerous countermeasures are proposed to improve CNNs' robustness against adversarial examples \cite{effi_defense,ob_grad,univ_at}, among which adversarial learning is considered as one of the most effective methods \cite{ob_grad}. Adversarial learning is essentially data augmentation with various adversarial examples \cite{FGSM,PGD,CW} to train a CNN. Madry \textit{et al.} \cite{PGD} first formulized adversarial learning into a saddle point problem, where the inner-maximization corresponds to the generation of adversarial examples while the outer-minimization is to achieve robust network parameters. However, the effectiveness of adversarial learning depends on the intensity of adversarial examples corresponding to the inner-maximization problem.

In practice, inner-maximization of adversarial learning requires multiple iterations to search more powerful perturbations, in which first-order gradient-based optimization is naturally adopted for its computing efficiency. However, it often lacks a detailed and global view of the landscape of optimization. Instead, second-order gradient-based methods can provide more information, including the trend of a first-order gradient, which helps to obtain more powerful adversarial examples \cite{SCORPIO}. Typically, second-order methods need calculate or approximate the Hessian matrix, especially in the context of exact deterministic optimization \cite{Fletcher,Lewis&Overton}. Since the requirement of quadratic storage and cubic computation time for each gradient update, second-order methods are not widely used in the most of machine learning tasks at present. The main challenge is to smooth the gap between its theoretical superiority and practical complexity. 

To further employ the advantage of second-order methods applied in the improvement of CNNs' robustness, we propose an novel approach to craft adversarial examples, named \textit{SOAE}s (Second-Order Adversarial Examples), for adversarial training. Our SOAE method constructs a new perturbation direction based on the second-order gradient information of the loss function in the Krylov subspace  \cite{FOM,krylov,GMRES}. Specifically, we use a Hessian-inverse to determine the adversarial direction, \textit{i.e.}, the direction of maximizing loss. However, directly solving the Hessian-inverse of a loss \textit{w.r.t} an image (in general with high resolution) requires quadratic storage and cubic operations. To address this problem, we approximate adversarial direction by a liner combination of Hessian-vector product in the Krylov subspace to reduce computation cost. Another benefit of optimization conducted in the Krylov subspace is to achieve an accurate approximation. We test SOAE's effectiveness on different models trained by various adversarial training techniques. Meanwhile, we use SOAEs to train several models and evaluate their robustness against various adversarial attack methods. Extensive experimental results demonstrate that our method achieves state-of-the-art performance in both attack and defense. Finally, we provide a deep insight into the effectiveness of our method from the theoretical perspective of computational complexity and attack-strength bound.

Our contribution is summarized as follows:
\begin{itemize}
	\item We propose a second-order gradient-based method to generate more powerful adversarial examples. Adversarial training through these adversarial examples enhance the models with higher robustness than those training with first-order adversarial examples.
	\item We address the problem of Hessian-inverse occurred in second-order gradients by transforming it into an optimization in the Krylov subspace, which remarkably reduce the computational complexity.
	\item We theoretically prove the priority of our method to PGD. Extensive experiments conducted on MINIST and CIFAR-10 also show our second-order method outperforms other methods including state-of-the-art AutoAttack.
\end{itemize}

\section{Related Work and Preliminary}

A clean example-label pair $(\boldsymbol{x},{y}) \sim \mathcal{D}$ is extracted from an underlying data distribution $\mathcal{D}$ in the standard classification task. The classifier ${f_{\theta} }( \cdot )$ with parameters $\boldsymbol{\theta}$ turns $\boldsymbol{x}$ into logits, namely the unnormalized probability values. The logits are normalized after a softmax layer and become a probability score. The softmax layer can be represented as a function ${p_k}(\boldsymbol{x}) = {e^{{f_{\theta ,k}}(\boldsymbol{x})}}/\sum\nolimits_l {{e^{{f_{\theta ,l}}(\boldsymbol{x})}}} $. Thus, the predicted class label is obtained by ${\hat y_k}(\boldsymbol{x}) = \arg {\max _k}{f_{\theta ,k}}(\boldsymbol{x})$. A standard training procedure is to minimize the empirical risk ERM \cite{ERM}). For the loss function $L(\boldsymbol{x},y,\boldsymbol{\theta} )$, the goal of standard training is formulized as
\begin{equation}
	\mathop {\min }\limits_\theta  {\mathbb{E}_{(x,y) \sim \mathcal{D}}}[L(\boldsymbol{x},y,\boldsymbol{\theta} )]
\end{equation}

Training the classifier through ERM principle guarantees a high accuracy on test sets but leads to an unavoidable vulnerability against adversarial attacks \cite{PGD}. To measure the immunity of a classifier ${f_\theta }( \cdot )$ against perturbations, adversarial robustness is defined with respect to a metric. In practice, the most used metric is an $L_p$-norm ($p=$ 1, 2, or $\infty$), combined with an $L_p$-ball ${B_p}(\epsilon ) = \{ \boldsymbol{\delta} | {\left\| \boldsymbol{\delta}  \right\|_p} \le \epsilon \} $. An adversarial example ${\boldsymbol{x}}^{adv}$ is obtained by adding perturbation $\boldsymbol{\delta}$ to the original example $\boldsymbol{x}$, where the upper bound of $\boldsymbol{\delta}$ is $\epsilon$. Practically, there are many adversarial example generation algorithms proposed to find ${\boldsymbol{x}^{adv}} = \boldsymbol{x} + \boldsymbol{\delta} $ such that $\boldsymbol{\delta}$ is very small but the model misclassifies $\boldsymbol{x}^{adv}$ to class label $\hat{y} \ne {y}$. Here $y$ is the ground-truth label of $\boldsymbol{x}$. A classifier is evaluated as robust to adversarial perturbation size $\epsilon$, if the label of the given input example $\boldsymbol{x}$ does not change for all perturbations of size up to $\epsilon$, \textit{i.e.}, $f_\theta(\boldsymbol{x}) = f_\theta({\boldsymbol{x}^{adv}}) = f_\theta(\boldsymbol{x} + \boldsymbol{\delta} )$ where $\boldsymbol{\delta}  \in {B_p}(\epsilon )$. In this scenario, $\epsilon$ is often called as a perturbation budget.

There are many algorithms to generate adversarial examples. Goodfellow \textit{et al.} \cite{FGSM} first proposed FGSM that multiplies the sign of loss gradient \textit{w.r.t.} the inputs to obtain the perturbation, \textit{i.e.}, $\boldsymbol{x}^{adv} =\boldsymbol{x} + \alpha \mathrm{sign} \nabla L(\boldsymbol{x},\boldsymbol{\theta})$. Madry \textit{et al.} \cite{PGD} further developed this single-iteration FGSM into a k-iteration method PGD with a projection step to restrict the size of perturbation, \textit{i.e.}, ${\boldsymbol{x}^{(t+1)}} = \mathrm{\Pi} ({\boldsymbol{x}^{(t)}} + \alpha \mathrm{sign}\nabla L(\boldsymbol{x},\boldsymbol{\theta}))$. Recently, Croce and Hein \cite{AutoAttack} proposed a reliable and stable attack method AutoAttack, which is an automatic parameter-free method integrating four attack methods: three white-box attacks APGD \cite{AutoAttack} with cross entropy loss, targeted APGD with difference-of-logits-ratio loss, and targeted FAB \cite{FAB}, and a black-box attack named SquareAttack \cite{SqAttack}. However, those methods mentioned above are all based on first-order gradient information. FGSM lacks accuracy due to its one-shot operation. Though PGD fixes this issue through k iterations, it undoubtedly prolongs the convergence time. Same situation occurs in AutoAttack since it uses the combination of several variant PGDs and other types of attacks. Besides, those first-order methods have some natural defects in approximating the loss landscape around the neighborhood of the input images.

Recently there are some second-order optimization methods applied to adversarial robustness. Li \textit{et al.} \cite{ASOD} proposed a new attack method based on an approximated second-order derivative of the loss function and showed this method can effectively reduce accuracy of adversarially trained models. However, there was still a noticeable gap between theoretical analysis and empirical results. Tsiligkaridis and Roberts \cite{SCORPIO} revealed that adversarial attack based on second-order approximation of the loss is more effective to models with regular landscape and decision boundaries. Ma \textit{et al.} \cite{SOAR} further studied the \textit{upper bound} and proposed SOAR, a second-order adversarial regularizer based on the Taylor approximation of the inner-max in the robust optimization objective. They showed that training with SOAR leads to significant improvement in adversarial robustness under $L_\infty$ and $L_2$ attacks. However, they also found its vulnerability to AutoAttack. A possible explanation is that SOAR overfits to a specific type of attack, \textit{i.e.}, loss function dependent.

The strictly limited application of the second-order methods in the field of adversarial robustness is mainly due to the high computation cost and the large storage consumption of the Hessian-related problem in the optimization process. Recently, a new second-order optimizer named Shampoo was proposed by Anil \textit{et al.} \cite{Shampoo}. With the efficiently utilized heterogeneous hardware architecture consisting of multi-core CPUs and multi-accelerator units, it outperformed the state-of-the-art first-order gradient descent methods in the field of image classification, large-scale machine translation, etc. The success of Shampoo undoubtedly reveals the potential for the development of second-order method in the deep learning tasks and encourage us to explore more possibilities of its application in adversarial robustness.

\section{Methodology}
\subsection{Second-Order Perturbation}

For a classifier ${f_{\theta} }( \cdot )$ and an input $\boldsymbol{x}$ with label $y$, our purpose is to find a tiny perturbation $\boldsymbol\delta$ that leads to a misclassification of ${f_\theta }( \cdot )$, \textit{i.e.}, 
\begin{equation}
	{f_\theta }({\boldsymbol{x}} + \boldsymbol\delta ) = {\hat{y}}, \quad s.t.\quad {y} \ne \hat{y} \land \boldsymbol\delta  \in {B_p}(\epsilon )
\end{equation}
Specifically, our goal is to maximize the loss function $L({\boldsymbol{x}} + \boldsymbol\delta ,y)$. Note that almost all the methods utilize the first-order gradient direction, \textit{i.e.}, $\partial L/ \partial \boldsymbol{x}$, to increase loss function. However, we think the second-order gradient including more global information for obtaining more powerful adversarial examples. For a clean example $\boldsymbol{x}$, the Taylor expansion of the loss function of its perturbated example $\boldsymbol{x}^{adv} = \boldsymbol{x} + \boldsymbol\delta$ can be written as:
\begin{equation}
	\label{eq3}
	L(\boldsymbol{x}^{adv}) \approx  Q(\boldsymbol\delta)
	= L(\boldsymbol{x}) + \nabla L{(\boldsymbol{x})^T}\boldsymbol\delta  + \frac{1}{2}{\boldsymbol\delta ^T}{\nabla ^2}L({\boldsymbol{x}})\boldsymbol\delta
\end{equation}
where the higher order terms are omitted.
This loss function is a quadratic function $Q(\boldsymbol\delta)$ of perturbation $\boldsymbol\delta$. Since we want the loss to increase as much as possible in each iteration, the final optimal perturbation can be presented by:
\begin{equation}
	\boldsymbol\delta^*  = \mathop {\arg\max}\limits_{\boldsymbol\delta  \in {B_p}(\epsilon )} Q(\boldsymbol\delta ).
\end{equation}
Set the derivative of $Q(\boldsymbol\delta)$ with respect to $\boldsymbol\delta$ to zero, we obtain a feasible optimization direction at the $t$-th iteration:
\begin{equation}
	\boldsymbol\delta^{(t)}  = {\left[ {{\nabla ^2}L({\boldsymbol{x}+\boldsymbol{\delta}^{(t-1)}})} \right]^{ - 1}}\nabla L({\boldsymbol{x}+\boldsymbol{\delta}^{(t-1)}})
\end{equation}
For convenience of expression, let ${\mathbf{H}} = {\nabla ^2}L(\cdot)$ and ${\mathbf{g}} = \nabla L(\cdot)$, then $\boldsymbol\delta^{(t)}  = {{\mathbf{H}}^{ - 1}}{\mathbf{g}}$. To limit the perturbation size, a step-size factor $\alpha$ is adopted. If the final perturbation is out of the limitation of the given $L_p$-ball, referring to the method in PGD, the final form of the generated adversarial example yields:
\begin{equation}
	\label{eq6}
	{\boldsymbol{x}^{adv}} =  \mathrm\Pi (\boldsymbol{x} + \alpha {{\mathbf{H}}^{ - 1}}{\mathbf{g}})
\end{equation}
where $\mathrm\Pi(\cdot)$ is a projection operator.

\subsection{Approximating 	${{\bf{H}}^{ - 1}}{\bf{g}}$}

To calculate the Hessian matrix requires quadratic storage and cubic operations, and obtaining its inverse in Equation \ref{eq6} is non-trivial. In this work, we approximate $\bf{H}^{- 1}\bf{g}$ in Equation \ref{eq6} by a liner combination of Hessian-vector product in the Krylov subspace to reduce computation cost:
\begin{equation}
	\label{eq7}
	\begin{split}
		\boldsymbol\delta^{(t)}&={{\bf{H}}^{ - 1}}{\bf{g}} \approx  \sum\limits_{i = 0}^{m - 1} {{\beta _i}{{\bf{H}}^i}{\bf{g}}}  \\
		&= {\beta _0}{\bf{g}} + {\beta _1}{\bf{Hg}} + {\beta _2}{{\bf{H}}^2}{\bf{g}} +  \cdots  + {\beta _{m - 1}}{{\bf{H}}^{m - 1}}{\bf{g}}
	\end{split}
\end{equation}
where $\beta_i$ are coefficients and $m$ is a hyperparameter ($m \ll \dim {\bf{H}}$). Here, an $m$-dimensional Krylov subspace ${\cal K}(\mathbf{H},\mathbf{g}) \triangleq \mathrm{span} \{\mathbf{g},\mathbf{Hg},\cdots,{\mathbf H}^{m-1}\mathbf{g}\}$.

We apply Generalized Minimum Residual (GMRES) \cite{GMRES} to solve Problem \ref{eq7}, which is one of the most effective orthogonalization method in the Krylov subspace. In an $m$-dimensional Krylov subspace ${\cal K}(\mathbf{H},\mathbf{g}) $, there exists a $\tilde{\boldsymbol{\delta}} \in \boldsymbol\delta^{(0)}+\cal K$ for a minimal residual, where $\boldsymbol{\boldsymbol\delta}^{(0)}$ is a randomly initialized perturbation. Hence, Problem \ref{eq7} is transformed into the following optimization to approximate $\boldsymbol\delta^{(t)}={{\bf{H}}^{ - 1}}{\bf{g}}$:
\begin{equation}
	\label{eq9}
	\boldsymbol\delta^{(t)}=\mathop {\arg\min}\limits_{\tilde{\boldsymbol\delta} \in {\boldsymbol\delta^{(0)}} + {{\cal K}}} {\left\| {\mathbf{g} - \mathbf{H}\tilde{\boldsymbol\delta}} \right\|_2}, \quad s.t. \quad
	(\mathbf{g} - \mathbf{H}\tilde{\boldsymbol\delta)} \bot \mathbf{H}{{\cal K}}	
\end{equation}
where  $\left\|\mathbf{g} - \mathbf{H}\tilde{\boldsymbol\delta}\right\|_2$ is the residual in affine space ${\boldsymbol\delta^{(0)}} + {{\cal K}}$, and $\mathbf{H}{{\cal K}}$ is a constraint space. However, directly solving this optimization is non-trivial. 
For any $ \tilde{\boldsymbol\delta} \in {\boldsymbol\delta^{(0)}} + {{{\cal K}}_m}$, there exists $\boldsymbol\gamma \in \mathbb{R}^m$ such that $\tilde{\boldsymbol{\delta}} = {\boldsymbol\delta^{(0)}} + {\bf{V}}_m\boldsymbol\gamma$, where ${\bf{V}}_m$ is an $m$-dimensional unitary matrix \cite{krylov}. Then we further transform Problem \ref{eq9} into an equivalent optimization problem. Let
\begin{equation}
	\label{eq0}
	\begin{split}
		&\mathbf{g} - \mathbf{H}\tilde{\boldsymbol{\delta}} = \mathbf{g} - \mathbf{H}({\boldsymbol\delta^{(0)}} + {\mathbf{V}}_m \boldsymbol\gamma) = {\mathbf{r}_0} - \mathbf{H}{\mathbf{V}_m}\boldsymbol\gamma \\
		&= \boldsymbol{\beta} {\mathbf{v}_1} - {\mathbf{V}_{m + 1}}{\mathbf{D}_{m + 1,m}}\boldsymbol\gamma = {\mathbf{V}_{m + 1}}(\boldsymbol{\beta} {\mathbf{e}_1} - {\mathbf{D}_{m + 1,m}}\boldsymbol{\gamma})
	\end{split}
\end{equation}
where ${\mathbf{r}_0} = \mathbf{g} - \mathbf{H}\boldsymbol\delta^{(0)}$, ${\mathbf{v}_i}$ is the $i$-th column vector of ${\mathbf{V}_m}$, ${\mathbf{D}_{m + 1,m}} = [d_{ij}]_{m+1 \times m}$ where $d_{ij}=({\mathbf{v}_i}, \mathbf{H}{\mathbf{v}_i})$, and ${\mathbf{e}_1} = {\left[ {1, 0, ..., 0} \right]^T} \in {\mathbb{R}^{m + 1}}$. Since the column vectors of the unitary matrix $\mathbf{V}_{m+1}$ are orthonormal, we have
\begin{equation}
	\begin{split}
		{\left\| {\mathbf{g} - \mathbf{H}\boldsymbol{\delta}} \right\|_2} &= {\left\| {{\mathbf{V}_{m + 1}}(\boldsymbol{\beta} {\mathbf{e}_1} - {\mathbf{D}_{m + 1,m}}\boldsymbol{\gamma)}} \right\|_2}\\
		&= {\left\| {\boldsymbol{\beta} {\mathbf{e}_1} - {\mathbf{D}_{m + 1,m}}\boldsymbol{\gamma}} \right\|_2}
	\end{split}
\end{equation}
Hence, Problem \ref{eq9} is turned into the following optimization:
\begin{equation}
	\label{eq11}
	\begin{split}
		\tilde{\boldsymbol{\delta}} &= {\boldsymbol\delta^{(0)}} + {\mathbf{V}_m} \boldsymbol\gamma^*\\
		\boldsymbol\gamma^*&=\mathop {\arg\min }\limits_{\boldsymbol\gamma \in \mathbb{R}{^m}} {\left\| {\boldsymbol\beta {\mathbf{e}_1} - {\mathbf{D}_{m + 1,m}}\boldsymbol\gamma} \right\|_2}
	\end{split}
\end{equation}

Since $m$ is sufficiently small in our work, we can further use QR-decomposition method to solve this least square problem. Let ${\mathbf{D}_{m + 1,m}} = \mathbf{Q}_{m + 1}^T{\mathbf{R}_{m + 1,m}}$ be the QR decomposition of ${\mathbf{D}_{m + 1,m}}$, where $\mathbf{Q}_{m + 1}^T \in {\mathbb{R}^{(m + 1) \times (m + 1)}}$ is an orthogonal matrix and ${\mathbf{R}_{m + 1,m}} \in {\mathbb{R}^{(m + 1) \times m}}$ is an upper triangle matrix. Then we have
\begin{equation}
	\label{eq12}
	\begin{split}
		{\left\| {\boldsymbol\beta {\mathbf{e}_1} - {\mathbf{D}_{m + 1,m}}\boldsymbol\gamma} \right\|_2} &= {\left\| {\boldsymbol\beta {\mathbf{e}_1} - \mathbf{Q}_{m + 1}^T{\mathbf{R}_{m + 1,m}}\boldsymbol\gamma} \right\|_2} \\
		& ={\left\| {\boldsymbol\beta \mathbf{Q}_{m + 1} {\mathbf{e}_1} - {\mathbf{R}_{m + 1,m}}\boldsymbol\gamma} \right\|_2} \\
		&= {\left\| {\boldsymbol\beta {\mathbf{q}_1} - \left[ {\begin{array}{*{20}{c}}
						{{\mathbf{R}_m}}\\
						0
				\end{array}} \right]\boldsymbol\gamma} \right\|_2}
	\end{split}
\end{equation}
where $\mathbf{q}_1$ is the first column of $\mathbf{Q}_{m + 1}$ and $\mathbf{R}_m$ represents the first $m$ rows of $\mathbf{R}_{m+1,m}$. Then, $\boldsymbol\gamma$ can be solved by the upper triangle equations
\begin{equation}
	\label{eq15}
	{\mathbf{R}_m} \boldsymbol\gamma = \boldsymbol\beta {\mathbf{q}_1}(1:m)
\end{equation}
where ${\mathbf{q}_1}(1:m)$ represents the vector consisting of the first $m$ elements of $\mathbf{q}_1$. Through solving these $m$-dimensional upper triangle equations by Numpy, the optimal $\boldsymbol\gamma^*$ is obtained. Substitute $\boldsymbol\gamma^*$ back into Equation \ref{eq11}, we finally obtain the optimal perturbation $\boldsymbol{\delta}^*$.

In practice, the QR decomposition of ${\mathbf{D}_{j + 1,j}}$ in Equation \ref{eq12} can be solved by a recursive Givens transformation method \cite{Givens}, \textit{i.e.}, obtain the QR decomposition of ${\mathbf{D}_{j + 1,j}}$ through a Givens transformation based on the QR decomposition of ${\mathbf{D}_{j - 1,j}}$. The QR decomposition of ${\mathbf{D}_{j,j - 1}} \in {\mathbb{R}^{j \times (j - 1)}}$ is presented as
\begin{equation}
	\begin{split}
			{\mathbf{D}_{j,j - 1}} &= {({\mathbf{G}_{j - 1}}{\mathbf{G}_{j - 2}}...{\mathbf{G}_1})^T}{\mathbf{R}_{j,j - 1}} \\
		&= \mathbf{Q}_j^T{\left[ {\begin{array}{*{20}{c}}
					{{\mathbf{R}_{j - 1}}}\\
					0
			\end{array}} \right]_{j \times (j - 1)}}
	\end{split}
\end{equation}
where ${\mathbf{R}_{j - 1}} \in {\mathbb{R}^{(j + 1) \times (j - 1)}}$ is an upper triangle matrix and $\mathbf{G}_i$ represents the Givens transformation. To ensure the consistency of matrix product, we suppose that the dimension of $\mathbf{G}_i$ will automatically increase according to our calculation requirement, \textit{i.e.}, expanding $\mathbf{G}_i$ by adding the identity matrix to the lower right corner. Briefly, we can obtain the last column of $\mathbf{R}_j$ by applying the Givens transformation ${\mathbf{G}_1},{\mathbf{G}_2},...{\mathbf{G}_j}$ to the last column of ${\mathbf{D}_{j + 1,j}}$ respectively.

\subsection{Second-Order Adversarial Examples}
\begin{algorithm}[t]
	\caption{Second-Order Adversarial Example (SOAE)}
	\label{alg:main}
	{\bfseries Input:} Original image $\boldsymbol{x}$, difference step-size $\eta>0$ and $\tau > 0$.\\
	{\bfseries Output:} $\boldsymbol{x}^{adv}=\boldsymbol{x}_N$
	\begin{algorithmic}[1]
		\FOR{$n = 1,2,\cdots,N$}
		\STATE $\mathbf{g} \leftarrow \nabla L({\boldsymbol{x}_n}), \boldsymbol{\delta}^{(0)} \leftarrow \mathbf{g}$
		\STATE ${\mathbf{Hg}} \leftarrow \left(\nabla L(\boldsymbol{x}_n + \eta \mathbf{g}) - \nabla L(\boldsymbol{x}_n)\right) / \eta $
		\STATE ${\mathbf{r}_0} \leftarrow \boldsymbol{\delta}^{(0)} - \mathbf{Hg}$, $\beta \leftarrow {\left\| {{\mathbf{r}_0}} \right\|_2}, {\mathbf{v}_1} \leftarrow {\mathbf{r}_0}/\beta $
		\FOR{$j=1,2,\ldots$}
		\STATE ${\mathbf{w}_j} \leftarrow \mathbf{H}{\mathbf{v}_j}$
		\FOR{$i=1$ {\bfseries to} $j$}
		\STATE ${h_{ij}} \leftarrow ({\mathbf{w}_j},{\mathbf{v}_i})$
		\STATE ${\mathbf{w}_j} \leftarrow {\mathbf{w}_j} - {h_{ij}}{\mathbf{v}_i}$
		\ENDFOR
		\STATE ${h_{j + 1,j}} \leftarrow {\left\| {{\mathbf{w}_j}} \right\|_2},{\mathbf{v}_{j + 1}} \leftarrow {\mathbf{w}_j}/{h_{j + 1,j}}$
		\IF{${\left\| \mathbf{\tilde r} \right\|_2}/\beta  < \tau $}
		\STATE $m \leftarrow j$, \textbf{break}
		\ENDIF
		\ENDFOR
		\STATE Solve ${\mathbf{R}_m}\boldsymbol\gamma^* = \beta {\mathbf{q}_1}(1:m)$
		\STATE $\boldsymbol{\tilde\delta} \leftarrow {\boldsymbol\delta^{(0)}} + {\mathbf{V}_m}\boldsymbol\gamma^*$
		\STATE ${\boldsymbol{x}_{n+1}} \leftarrow  \mathrm{Clip}\{{\boldsymbol{x}_n} + \frac{\alpha}{N}\boldsymbol{\tilde\delta}\}$
		\ENDFOR
	\end{algorithmic}
\end{algorithm}

We illustrate the entire procedure of generating our second-order adversarial example (SOAE) in Algorithm \ref{alg:main}. Our goal is to construct a second-order adversarial perturbation $\boldsymbol{\delta}^*=\mathbf{H}^{-1}\mathbf{g}$ and output the final adversarial example $\boldsymbol{x}^{adv}=\boldsymbol{x}+\alpha\boldsymbol{\delta}^*$, which is implemented by multi-iterations to approximate $\boldsymbol{\delta}^*$ in our algorithm. To gain a better performance, the step size is divided into $\alpha / N$ in each iteration, where $N$ is the number of iterations. If the calculated perturbation is larger than the perturbation budget $\epsilon$, we apply a projection step like PGD \cite{PGD}. To avoid directly computing the Hessian matrix, we apply a difference approximation ${\mathbf{Hg}} = (\nabla L(\boldsymbol{x} + \eta \mathbf{g}) - \nabla L(\boldsymbol{x}))/\eta $ in step 3, which is proved to be an accurate approximation to the Hessian-vector product \cite{SCORPIO}. Moreover, it has been proved that gradient direction is well aligned with the direction of maximum curvature of models \cite{curv1,curv2}, which makes $\boldsymbol{\delta}^{(0)} \leftarrow \mathbf{g}$ a strong initialization for our algorithm. The inner loop of our algorithm (step 5 to 15) is the Hessian inverse approximation, which controls the dimension of the Krylov subspace through a threshold $\tau$. In practice, the dimension $m$ can be compressed to a sufficiently small size that is $1/20$ of the original Hessian's dimension, which keep the approximation's exactness while significantly reduce the computation cost. Our algorithm involves two hyperparameters: 1) the difference step-size $\eta$ to control the accuracy of the difference approximation and 2) the approximation threshold $\tau$ to determine the approximating precision of Hessian inverse.

\subsection{Second-Order Adversarial Training}

We use our SOAEs to attack the models under various adversarial training such as PGD \cite{PGD} and TRADES \cite{trades}. Experimental results in Section \ref{sec:exp} show that our SOAEs can deceive the first-order adversarial training models with high success rates. This inspires us to use our SOAEs to train models, which is referred to as second-order adversarial training (SOAT) in this paper. The optimization of second-order adversarial training is formulized as
\begin{equation}
	\mathop {\min }\limits_{\boldsymbol\theta} L(\boldsymbol\theta;\boldsymbol{x}+\boldsymbol\delta^*,y)
	\quad s.t. \quad \boldsymbol\delta^* = {\boldsymbol\delta^{(0)}} + {\mathbf{V}_m}\boldsymbol\gamma^*
\end{equation}
where $L$ is a loss function and $\boldsymbol\delta^*$ is the second-order perturbation of $\boldsymbol{x}$. After the second-order adversarial training, the model gain the robustness against both first-order and second-order attacks.

\subsection{Theoretical Analysis on Attack Strength}
Recall in Equation \ref{eq3} that we consider the loss function $L(\boldsymbol{x+\delta})$ as a quadratic function $Q(\boldsymbol{\delta})$ with respect to our second-order perturbation $\boldsymbol{\delta}$ and the final perturbation is obtained through the following optimization:
\begin{equation}
	\begin{split}
		Q(\boldsymbol\delta)
		= L(&\boldsymbol{x}) + \nabla L{(\boldsymbol{x})^T}\boldsymbol\delta  + \frac{1}{2}{\boldsymbol\delta ^T}{\nabla ^2}L({\boldsymbol{x}})\boldsymbol\delta \\
		&\boldsymbol{\delta}^* = \mathop {\arg\max}\limits_{\boldsymbol\delta  \in {B_p}(\epsilon )} Q(\boldsymbol\delta )
	\end{split}
\end{equation}
Note that in the construction of our second-order perturbation $\boldsymbol{\delta}^*$, we use a quadratic function $Q(\boldsymbol\delta)$ instead of maximizing the original loss $L(\boldsymbol{x}+\boldsymbol\delta)$. Assume that there exists an optimal attack $\hat{\boldsymbol{\delta}}$ which directly comes from maximizing $\hat L = L(\boldsymbol{x}+\hat{\boldsymbol{\delta}})$. In such case, 
\begin{equation}
	\begin{split}
		\hat L - Q &= L(\boldsymbol{x}+\hat{\boldsymbol{\delta}})-L(\boldsymbol{x})-(L(\boldsymbol{x}+\boldsymbol\delta^*)-L(\boldsymbol{x}))\\ &= Q(\hat{\boldsymbol{\delta}})-Q(\boldsymbol\delta^*)+o(\boldsymbol{x}^3)
	\end{split}	
\end{equation}
where $\hat{\boldsymbol{\delta}}$ and $\boldsymbol\delta^*$ represent the ``optimal" perturbation and our second-order perturbation respectively and $o(\boldsymbol{x}^3)$ is the three-order Taylor's polynomials. Hence, 
\begin{equation}
	\label{eq18}
	\begin{split}
		Q(\hat{\boldsymbol{\delta}})-Q(\boldsymbol\delta^*)&=\mathbf{g}^T\hat{\boldsymbol{\delta}}+\frac{1}{2}\hat{\boldsymbol{\delta}}^T\mathbf{H}\hat{\boldsymbol{\delta}}-(\mathbf{g}^T\boldsymbol\delta^*+\frac{1}{2}\boldsymbol\delta^{*T}\mathbf{H}\boldsymbol\delta^*)
		\\&=(\hat{\boldsymbol{\delta}}-\boldsymbol\delta^*,\mathbf{g})+(\hat{\boldsymbol{\delta}}-\boldsymbol\delta^*,\frac{1}{2}\mathbf{H}(\hat{\boldsymbol{\delta}}+\boldsymbol\delta^*))	
	\end{split}
\end{equation}
By applying Holder inequality to the last term of Equation \ref{eq18},
\begin{equation}
	\begin{split}
		|(\hat{\boldsymbol{\delta}}-\boldsymbol\delta^*,\frac{1}{2}\mathbf{H}(\hat{\boldsymbol{\delta}}+\boldsymbol\delta^*))|&\le\frac{1}{2}\|\hat{\boldsymbol{\delta}}-\boldsymbol\delta^*\|_p \cdot (\|\mathbf{H}\hat{\boldsymbol{\delta}}\|_q+\|\mathbf{H}\boldsymbol\delta^*\|_q)\\ &\le \epsilon^2\left\|\mathbf{H}\right\|_{p,q}
	\end{split}
\end{equation}
where $\epsilon$ is a perturbation budget. Hence, the bound of real adversarial loss and our second-order loss yields:
\begin{equation}
	|\hat L-Q| \le 2\epsilon\left\|\mathbf{g}\right\|_q+2\epsilon^2\left\|\mathbf{H}\right\|_{p,q}+\frac{\epsilon^3M}{3}
\end{equation}
where $M$ is the supremum-norm of all derivatives of $L(\boldsymbol{x})$ of
order three. This inequality reveals that the proximity of our second-order adversarial perturbation to the optimal perturbation is controlled by the attack strength and the network regularity. A greater regularity of the network will result in a smaller gap between our attack and the optimal attack. It has been experimentally demonstrated that various robust methods, such as adversarial training and geometric regularization lead models to a more regular loss landscape than their non-robust counterparts \cite{Lyu,Ros,Moo}. Therefore, our SOAE attack is effective to various adversarially trained models, which is experimentally demonstrated in Section \ref{sec:exp}.

\subsection{Time Complexity Analysis}

For convenience of theoretical analysis, we take a three-layer fully connected (FC) neural network as an example, since the construction of adversarial examples contains multiple forward and backward propagation. Our FC network contains $l_1$, $l_2$, and $l_3$ neurons in the input, hidden and output layer respectively, where $l_2 > l_1$ and $l_2 > l_3$. With one original image resized to $(l_1, 1)$, the forward propagation requires two multiplication of weight matrix and activation vector, that is, $l_1 \times l_2 + l_2 \times l_3$ calculation. Its time complexity is $\mathcal{O}(l_1l_2+l_2l_3)$. Similarly, the time complexity of back propagation is $\mathcal{O}(l_1l_2+l_2l_3)$ as well. Hence, the gradient calculation for SOAE's initialization has $\mathcal{O}(l_2^2)$ time complexity.

Our calculated gradient $\mathbf{g}$ has the same dimension as the input, \textit{i.e.}, $\dim\mathbf{g}=l_1$. Step 3 and Step 4 only contains multiplication of numbers and $l_1$-dimensional vectors. Thus, its complexity is $\mathcal{O}(l_1)$. Note that Step 6 does not require Hessian $\mathbf{H}$, instead, we have
\begin{equation}
	\begin{split}
		\mathbf{H}\mathbf{v}_1&=\frac{1}{\beta}\mathbf{H}(\boldsymbol{\delta}^{(0)}-\mathbf{Hg})=\frac{1}{\beta}(\mathbf{Hg}-\mathbf{HHg})\\
		&=\frac{1}{\beta}[\frac{\nabla L(\boldsymbol{x}_n + \eta \mathbf{g}) - \nabla L(\boldsymbol{x}_n)}{\eta}\\
		&-\frac{\nabla L(\boldsymbol{x}_n + \eta \mathbf{Hg}) - \nabla L(\boldsymbol{x}_n)}{\eta}]
	\end{split}
\end{equation}
where the Hessian-vector product $\mathbf{Hg}$ is approximated through finite difference:
\begin{equation}
	{\mathbf{Hg}} = \left(\nabla L(\boldsymbol{x}_n + \eta \mathbf{g}) - \nabla L(\boldsymbol{x}_n)\right) / \eta 
\end{equation}
where the same finite difference can be applied to $\mathbf{HHg}$ by consider it as another Hessian-vector product, in which case the corresponding vector is the product $\mathbf{Hg}$, \textit{i.e.},
\begin{equation}
	\mathbf{H}^2\mathbf{g} = \left(\nabla L(\boldsymbol{x}_n + \eta \mathbf{Hg}) - \nabla L(\boldsymbol{x}_n)\right) / \eta .
\end{equation}

Similarly, all $\mathbf{w}_j=\mathbf{Hv}_j$ can be calculated through
\begin{equation}
	\mathbf{w}_j=\mathbf{Hv}_j=\mathbf{H}(\mathbf{Hv}_{j-1})=\cdots=
	\mathbf{H}(\mathbf{H}\cdots(\mathbf{H}(\mathbf{Hv}_1))).
\end{equation}
In $j$ loops where $j$ is up to $m$, each calculation of $\mathbf{w}_j$
requires once forward and once backward propagation. Its time complexity is $\mathcal{O}(ml_2^2)$.

The $i$ loop (Step 7 to 10) only contains inner product of $l_1$-dimensional vectors and number-vector multiplication. The time complexity of one loop is $\mathcal{O}(l_1)$. Note that $i$ is up to $j$ where $j$ is up to $m$, the total time complexity of the $i$ loop yields $\mathcal{O}(l_1(1+2+\cdots+m))=\mathcal{O}(m^2l_1)$.

After the $j$ loop there is an $m$-dimensional upper triangle linear equations in Step 16 which comes from the QR decomposition of ${\mathbf{D}_{m + 1,m}} = [d_{ij}]_{m+1 \times m}$ where $d_{ij}=({\mathbf{v}_i}, \mathbf{H}{\mathbf{v}_i})$, referring to Eq. \ref{eq0}. The time complexity of the QR decomposition of an m-dimensional matrix is $\mathcal{O}(m^3)$ when applying Givens transformation \cite{Givens}. Solving the upper triangle equations in Step 16 has $\mathcal{O}(m^2)$ time complexity. The rest operations followed by Step 16 are some simple operations on $l_1$-dimensional vectors which have $\mathcal{O}(l_1)$ complexity.

Now we can give a formal estimate of our SOAE. For a three-layer FC network with input size $(l_1,1)$, hidden layer with $l_2$ neurons, the time complexity of SOAE is
\begin{equation}
	\label{eqtc}
	\mathcal{O}(l_2^2+ml_2^2+m^2l_1+m^2).
\end{equation}
Here, $m$ is controlled by a threshold $\tau$ in Step 12, which we will give a detailed discussion in Sec. \ref{sec:hyper}. In practice, $m$ is relevant with input size $l_1$ and a small $m$ value is sufficient enough to guarantee a high performance. This theoretical analysis reveals SOAE's effectiveness on small scale datasets. Even for large scale images such as ImageNet, our method still shows competitive performance in practical experiments.

In addition, there is no extra storage requirement for our algorithm since it is Hessian-free, which means the batch size can be set as the same as other attack methods. Such design makes our method more competitive than current second-order adversarial methods.

\begin{table*}[t]
	\caption{Experimental Results on MNIST. STD represents standard training with clean examples.}
	\label{tab:mnist}
	\begin{center}
		\begin{small}
			\begin{tabular}{cccccccc}
				\toprule
				Models   & Training & Clean    & FGSM    & PGD    & Auto   & ASOD    & \textbf{SOAE}    \\
				Used     & Methods  & Examples & Attack  & Attack & Attack & Attack & Attack  \\
				\midrule
				\multirow{7}{*}{ResNet18}
				&STD  & 97.78\%  & 0.07\%  &  0.06\%&  0.01\%&  0.01\%&  0.01\% \\
				~        &PGD \cite{PGD}      & 86.34\%  & 63.40\% & 55.89\%& 49.10\%& 20.20\%& 30.85\% \\
				~        &TRADES \cite{trades}   & 85.98\%  & 65.05\% & 53.87\%& 40.27\%& 39.76\%& \textbf{38.32}\% \\
				~        &SOAR \cite{SOAR}     & 87.95\%  & 67.15\% & 56.06\%& 18.25\%& 29.12\%& 20.14\% \\
				~        &STN \cite{ASOD}      & 86.26\%  & 66.77\% & 54.90\%& 29.81\%& 48.74\%& \textbf{25.25}\% \\
				~        & SCORPIO \cite{SCORPIO} & 87.92\% &68.18\% &58.44\%&41.01\%&47.62\%&\textbf{39.03}\%\\
				~        &\textbf{SOAT}      & 87.62\%  & \textbf{70.89}\% & \textbf{60.13}\%& 39.10\%& \textbf{49.45}\%& \textbf{47.53}\% \\
				\midrule
				\multirow{6}{*}{WideResNet}
				&STD  & 99.90\%  &  0.08\% &  0.05\%&  0.01\%&  0.02\%&  0.01\% \\
				~        &PGD \cite{PGD}      & 84.55\%  & 66.50\% & 57.12\%& 44.51\%& 21.76\%& 29.84\% \\
				~        &TRADES \cite{trades}   & 80.56\%  & 68.32\% & 56.49\%& 42.79\%& 41.37\%& \textbf{32.15}\% \\
				~        &SOAR \cite{SOAR}     & 86.99\%  & 68.64\% & 57.80\%& 19.13\%& 33.90\%& 27.60\% \\
				~        &STN \cite{ASOD}      & 87.42\%  & 66.81\% & 53.21\%& 33.91\%& 46.25\%& \textbf{30.45}\% \\
				~        & SCORPIO \cite{SCORPIO} & 89.01\% &67.43\% & 55.02\%&39.88\%&43.69\%&\textbf{38.24}\%\\
				~        &\textbf{SOAT}      & 85.23\%  & \textbf{69.98}\% & 57.73\%& \textbf{45.22}\%& 45.96\%& \textbf{43.47}\% \\
				\bottomrule
			\end{tabular}
		\end{small}
	\end{center}
\end{table*}

\section{Experiment}
\label{sec:exp}
In this section, extensive experiments are conducted to evaluate the effectiveness of our SOAE and SOAT. We train ResNet18 and WideResNet using various methods including standard training and adversarial training on MNIST and CIFAR-10 and test them with various attack methods. For large-scale image evaluation, we use 100,000 samples of 100 classes from ImageNet to construct our own dataset. After the effectiveness evaluation, we analyze the hyperparameters in our methods to find the most proper values.

\subsection{Experimental Setup}

\textbf{Datasets:} MNIST contains 60,000 training examples and 10,000 test examples of handwriting numbers with the size of 28$\times$28. CIFAR-10 contains 60,000 RGB images of 10 classes with the size of 32$\times$32, which are divided into a training set containing 50,000 examples and a test set containing 10,000 examples. ILSVRC2012 is a subset of ImageNet containing 1.2 million 224$\times$224 images of 1,000 classes, where we randomly select 100,000 images of 100 classes to construct our own dataset ImageNet-100 for large-scale image evaluation.

\textbf{Adversarial Attack:} We use five different methods to generate adversarial examples: (1) FGSM \cite{FGSM} with the perturbation size $\epsilon = 8 / 255$; (2) a 20-iteration PGD \cite{PGD} with the step size $2/255$ and the total perturbation size $\epsilon = 8 / 255$; (3) AutoAttack \cite{AutoAttack} with L$_\infty$-norm and $\epsilon = 8 / 255$; (4) ASOD \cite{ASOD} with the total perturbation size $\epsilon = 8 / 255$; and (5) our SOAE with the total perturbation size $\epsilon = 8 / 255$.

\textbf{Adversarial Training:} We train ResNet18 and WideResNet with two first-order adversarial training methods: (1) PGD with 7 iterations and the step size is $2/255$, (2) TRADES with the total perturbation $\epsilon=8/255$ and the step size is $2/255$; three second-order adversarial training methods: (1) SOAR \cite{SOAR} in which we follow the original settings, (2) STN \cite{ASOD} trained with ASOD, (3) SCORPIO regularizer \cite{SCORPIO}, and (4) our method SOAT. All training phases have 300 epochs with the batch size of 128. We use SGD with a momentum of 0.9, where the initial learning rate is 0.1 with a weight decay of 0.0003.

\subsection{Results on MNIST and CIFAR10}

\begin{table*}[t]
	\caption{Experimental Results on CIFAR-10.}
	\label{tab:cifar}
	\begin{center}
		\begin{small}
			\begin{tabular}{cccccccc}
				\toprule
				Models   & Training & Clean    & FGSM    & PGD    & Auto   & ASOD    & \textbf{SOAE}    \\
				Used     & Methods  & Examples & Attack  & Attack & Attack & Attack & Attack  \\
				\midrule
				\multirow{7}{*}{ResNet18}
				&STD  & 92.54\%  & 30.59\% & 0.14\% &  0.05\%& 0.12\% & 0.08\%  \\
				~        &PGD \cite{PGD}      & 80.64\%  & 50.96\% & 42.86\%& 40.59\%& 40.32\%& \textbf{38.19}\% \\
				~        &TRADES \cite{trades}   & 85.61\%  & 53.06\% & 45.38\%& 41.34\%& 43.49\%& \textbf{40.26}\% \\
				~        &SOAR \cite{SOAR}     & 87.95\%  & 55.70\% & 56.06\%& 19.66\%& 45.08\%& 35.45\% \\
				~        &STN \cite{ASOD}      & 83.03\%  & 54.83\% & 52.34\%& 38.00\%& 48.15\%& \textbf{37.98}\% \\
				~        & SCORPIO \cite{SCORPIO} & 86.58\% &55.07\%  & 49.56\% & 39.27\%   & 46.82\% & \textbf{38.75}\%\\
				~        &\textbf{SOAT }     & 85.47\%  & \textbf{56.24}\% & 54.72\%& \textbf{45.98}\%& 47.73\%& \textbf{49.27}\% \\
				\midrule
				\multirow{7}{*}{WideResNet}
				&STD  & 93.79\%  & 44.77\% & 0.03\% & 0.01\% & 0.13\% & 0.09\% \\
				~        &PGD \cite{PGD}      & 81.83\%  & 51.15\% & 43.66\%& 42.62\%& 41.86\%& \textbf{39.14}\% \\
				~        &TRADES \cite{trades}   & 86.54\%  & 54.66\% & 45.70\%& 43.70\%& 44.22\%& \textbf{41.21}\% \\
				~        &SOAR \cite{SOAR}     & 88.02\%  & 67.15\% & 57.92\%& 20.41\%& 47.31\%& 39.35\% \\
				~        &STN \cite{ASOD}      & 84.11\%  & 58.59\% & 53.16\%& 38.95\%& 49.27\%& \textbf{38.02}\% \\
				~        & SCORPIO \cite{SCORPIO} & 86.99\% &61.37\% & 50.50\% &41.28\%&47.33\%& \textbf{40.11}\%\\
				~        &\textbf{SOAT}      & 86.38\%  & 62.33\% & 57.67\%& \textbf{47.53}\%& \textbf{50.50}\%& \textbf{51.46}\% \\
				\bottomrule
			\end{tabular}
		\end{small}
	\end{center}
\end{table*}

\textbf{Results on MNIST:}  The experimental results in Table \ref{tab:mnist} and Table \ref{tab:cifar} illustrate the effectiveness of our methods. Compared with the first-order adversarial attacks, our SOAE make the models' accuracy decrease remarkably, though these models are adversarially trained. Compared with the second-order ASOD attack, our SOAE attack still achieves equivalent or larger accuracy drop. Nevertheless, ASOD \cite{ASOD} is merely effective to compromise PGD adversarial training, our SOAE can disrupt a wide range of adversarial training techniques. On the other hand, if the model is adversarially trained with SOAEs, named SOAT, it can significantly improve the network robustness, particularly resistant to second-order adversarial attacks. Even under the strongest AutoAttack, SOAT still shows state-of-the-art performance.

\textbf{Results on CIFAR-10:} We also test our method on the CIFAR-10 dataset. Experimental results as well illustrate the efficiency of our method. Both of our attack (SOAE) and defense methods (SOAT) achieve state-of-the-art performance. Except for AutoAttack on the SOAR-trained models, which is claimed in \cite{SOAR} that their method is highly vulnerable to AutoAttack, our SOAE outperforms all the rest attacks. On the other hand, the models trained by SOAT still guarantee high robustness against various adversarial attacks. It even achieves the highest accuracy on two trained models under AutoAttack.

\subsection{Results on ImageNet-100}

We test the effectiveness of our algorithm on our large-scale image dataset ImageNet-100. As analyzed in previous section, SOAE's speed is related to input size. Large scale images may theoretically increase SOAE's convergence time. This effect is not obvious on MNIST and CIFAR-10 that since the approximation dimension $m$ is small when dealing with 28$\times$28 and 32$\times$32 images ($m=15$ will be sufficient enough, as shown in Sec. \ref{sec:hyper}). However, when encountering 224$\times$224 images, the required $m$ value unavoidably increases to more than 100. Such situation makes $m$'s effect significant in Eq. \ref{eqtc}. However, additional running time is proved to be worthy according to experimental results. Our SOAE attack outperforms other attack methods in all situations. It successfully reduces more than 8\% accuracy compared with AutoAttack on a PGD-training model. On four different adversarial training models, its performances are all better than other attack methods. Training with SOAE also brings us satisfying results, where our SOAT model has the highest accuracy under all three attacks.

\begin{table}[t]
	\caption{Experimental Results of ResNet18 on ImageNet-100.}
	\label{tab:imagenet}
	\begin{center}
		\begin{small}
			\begin{tabular}{ccccc}
				\toprule
				Training & Clean & PGD  & Auto  & \textbf{SOAE}    \\
				Methods  & Examples& Attack & Attack & Attack  \\
				\midrule
				STD  & 91.33\%  & 0.02\% &  0  &  0 \\
				PGD  & 78.26\% & 55.37\%& 48.66\%& 40.31\% \\
				TRADES   & 82.91\%  & 56.09\%& 47.25\%& 40.05\% \\
				SOAR & 80.80\%  & 61.68\%&  45.12\%&  42.35\%\\
				\textbf{SOAT} & 84.50\%  & 62.84\%& 52.99\%&  47.90\% \\
				\bottomrule
			\end{tabular}
		\end{small}
	\end{center}
\end{table}

\subsection{Hyperparameter Analysis}
\label{sec:hyper}

In this subsection, we mainly discuss three hyperparameters in our method: the difference step-size $\eta$, approximation threshold $\tau$, and the iteration time $N$.

\begin{figure}[h]
	\centering
	\includegraphics[width=0.55\columnwidth]{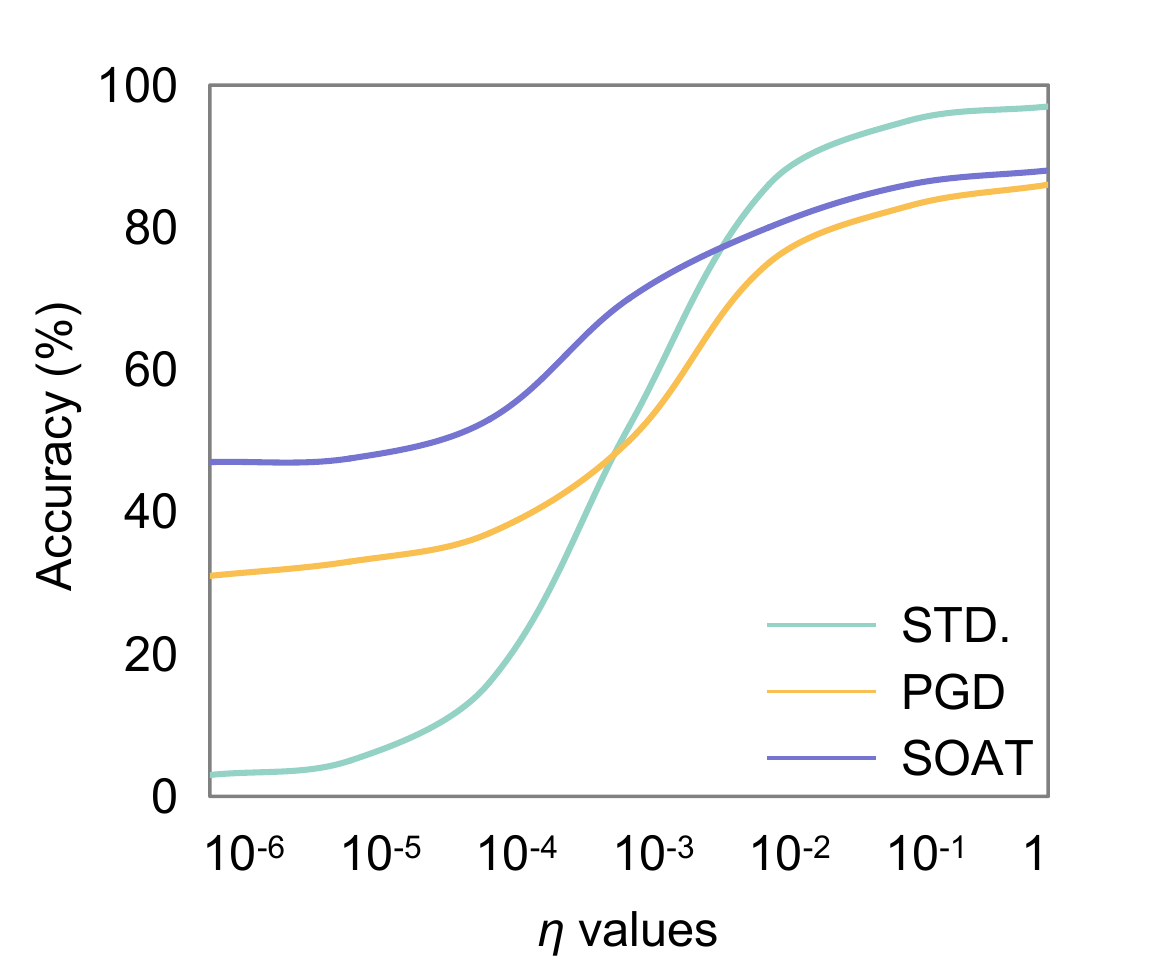}
	\caption{Accuracy change under different $\eta$ values. STD, PGD, and SOAT represent different training methods.}
	\label{fig:eta}
\end{figure}

\textbf{Difference Step-size $\eta$:} Recall that in step 3 of Algorithm \ref{alg:main}, we use a first-order difference of gradients to approximate the Hessian-vector product and the step-size $\eta$ controls the accuracy of this approximation. We evaluate how different $\eta$ values influence the performance of our algorithm. Specifically, we generate adversarial examples on MNIST with different $\eta$ values and use them to test the accuracy of the standard-trained, PGD-trained and SOAT-trained ResNet18 models respectively as shown in Figure \ref{fig:eta}. Our SOAE attack achieves the best performance as $\eta$ is less than $10^{-5}$. But if $\eta$ is larger than $10^{-4}$, it bring an obvious ineffectiveness to our method. As $\eta$ increases to $10^{-2}$, most perturbed examples generated by our algorithm are actually not adversarial examples. It means that a large $\eta$ will cause an extremely poor approximation to the Hessian-vector product and lead the subsequent step of the algorithm to a complete blindness. In conclusion, we suggest the proper $\eta$ value should be no more than $10^{-5}$.

\begin{figure}[h]
	\centering
	\includegraphics[width=0.6\columnwidth]{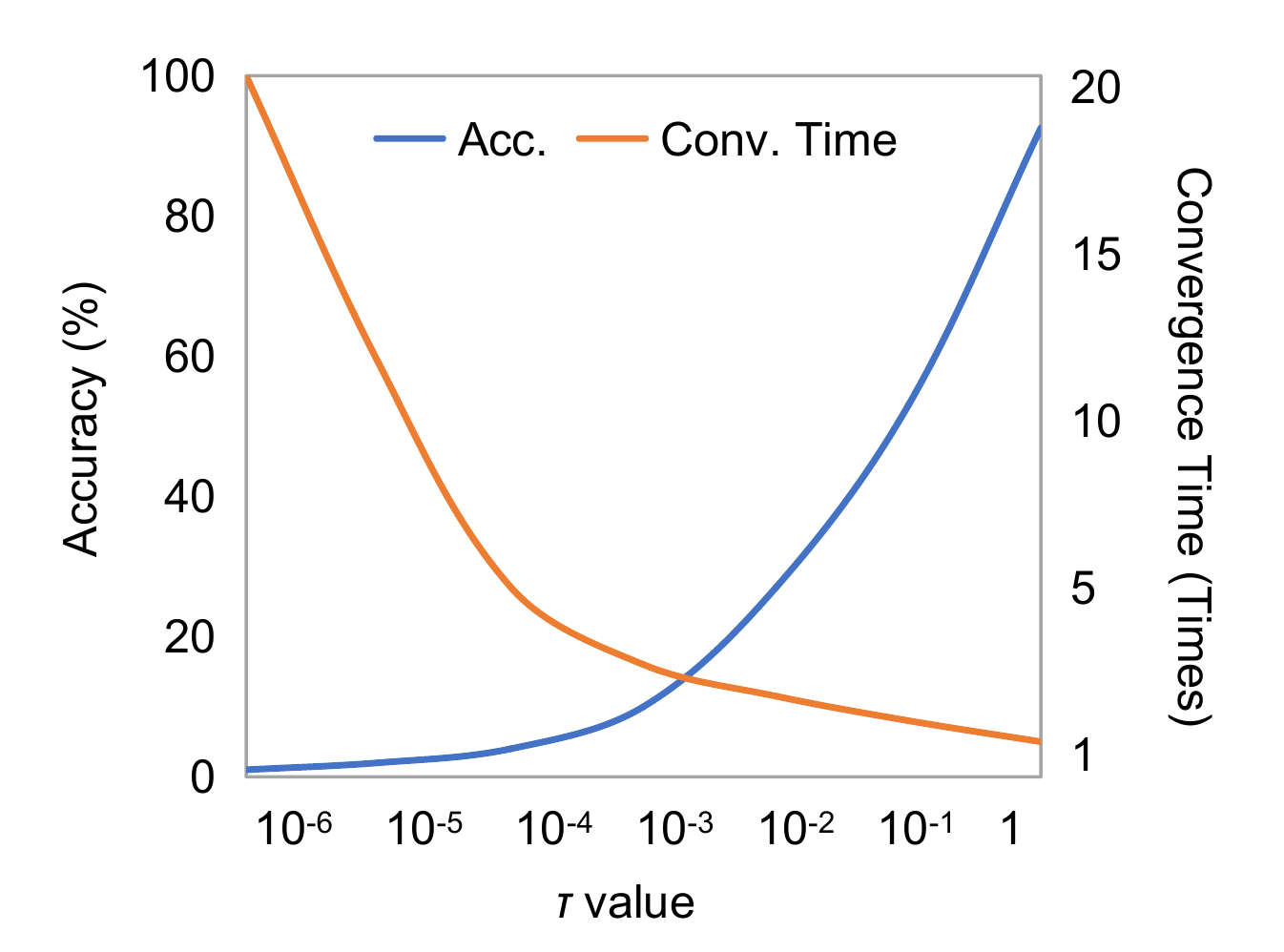}
	\caption{Accuracy change of ResNet18 and convergence time under different $\tau$ values. The convergence time is multiples of the time when $\tau = 1$.}
	\label{fig:tau}
\end{figure}

\textbf{Approximation Threshold $\tau$:} This hyperparameter controls the approximating precision of the Hessian-inverse. Intuitively, a smaller $\tau$ leads to a more accurate approximation. However, too small a $\tau$ will significantly increases the convergence time. As is analyzed in previous section, $\tau$ value determines the approximation dimension $m$, which significantly influences our algorithm's running time. We use a series of $\tau$ values to respectively generate 1,000 adversarial examples on CIFAR-10 dataset and attack the standard-trained ResNet18 model. The accuracy change and the convergence time are shown in Figure \ref{fig:tau}. We notice that when $\tau < 10^{-3}$, our SOAE can achieve good performance, where the accuracy drops below 16\%. We further find that the accuracy barely changes from $\tau = 10^{-6}$ to $10^{-3}$, however, the convergence time drops significantly. Therefore, we suggest that $10^{-4} \le \tau \le 10^{-3}$ is proper in practice. We also record the $m$ values corresponding to different $\tau$ values on different datasets, as shown in Tab. \ref{tab:m}

\begin{table}[t]
	\caption{Approximation dimension $m$ under different $\tau$ values.}
	\label{tab:m}
	\begin{center}
		\begin{small}
			\begin{tabular}{cccccc}
				\toprule
				\multirow{2}{*}{Dataset}& \multicolumn{5}{c}{$\tau$ value}\\
				~  &$10^{-5}$&$10^{-4}$&$10^{-3}$ &$10^{-2}$&$10^{-1}$\\
				\midrule
				MNIST(28$\times$28)&152&71&43&26&17\\
				CIFAR-10(32$\times$32)&243&117&65&38&22\\
				ImageNet(224$\times$224)&2196&1324&746&327&221\\
				\bottomrule
			\end{tabular}
		\end{small}
	\end{center}
\end{table}

\begin{figure}[h]
	\centering
	\includegraphics[width=0.55\columnwidth]{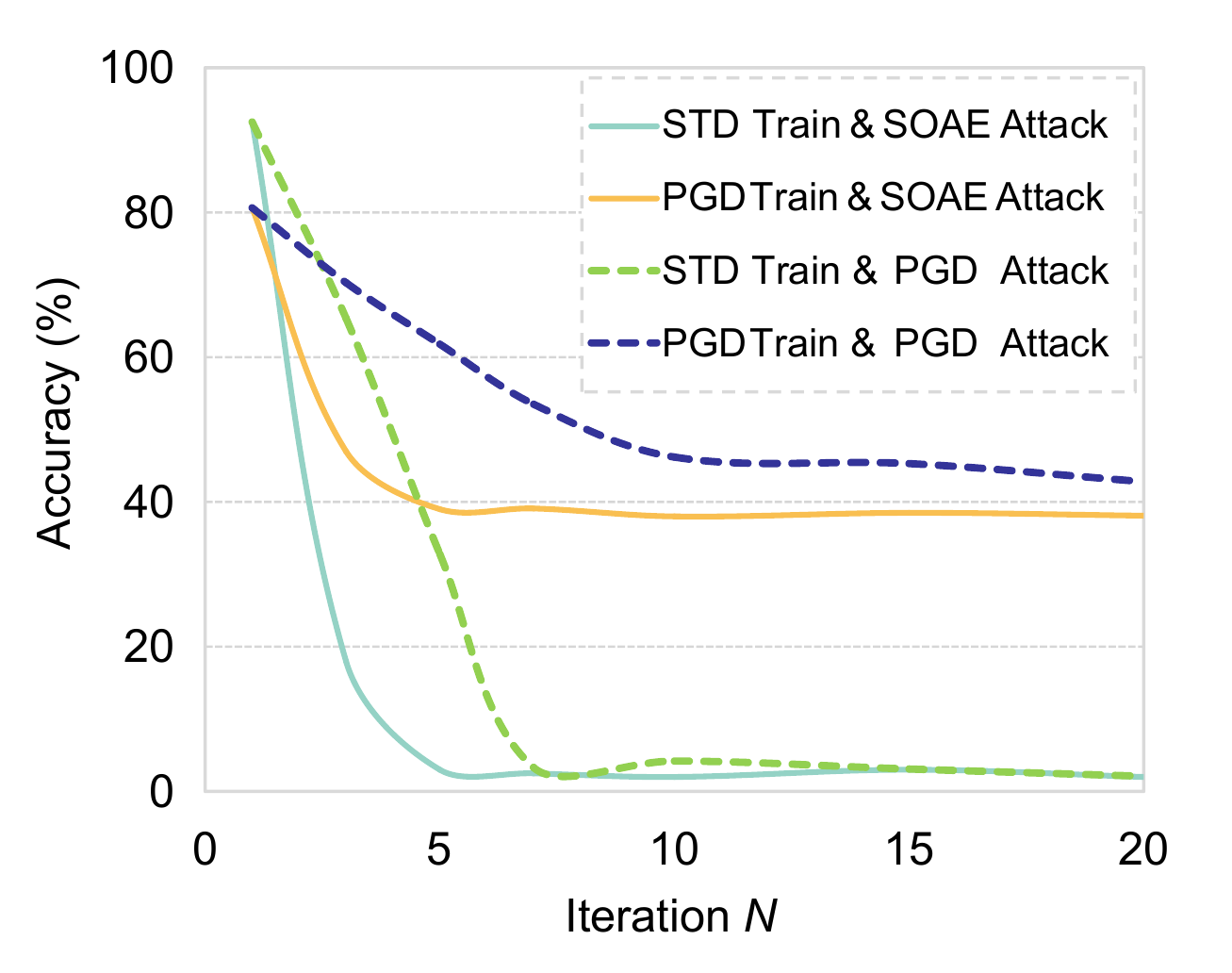}
	\caption{Accuracy change of ResNet18 model with the increasing of iteration $N$. We run PGD and SOAE for 20 iterations respectively and train the model with the standard training and PGD training.}
	\label{fig:N}
\end{figure}

\textbf{Total Iteration $N$:} The total iterations have a direct influence on attack success rates. But a large number of iteration will lead to insufferable time-consuming. In order to find proper iterations to achieve a good trade-off between attack success rates and computation costs, we run 20 iterations for our algorithm and PGD respectively on ResNet18 and record the accuracy changes  every iteration in Figure \ref{fig:N}. Compared with PGD using 7 or 20 iterations to converge, our method need only less than 4 iterations but yielding a competitive performance. This further proves that our second-order method has a perfect convergence property. We believe that our method is practical and it will be faster after further optimization.


\subsection{Visualization}

\begin{figure}[h]
	\begin{center}
		\includegraphics[width=0.9\columnwidth]{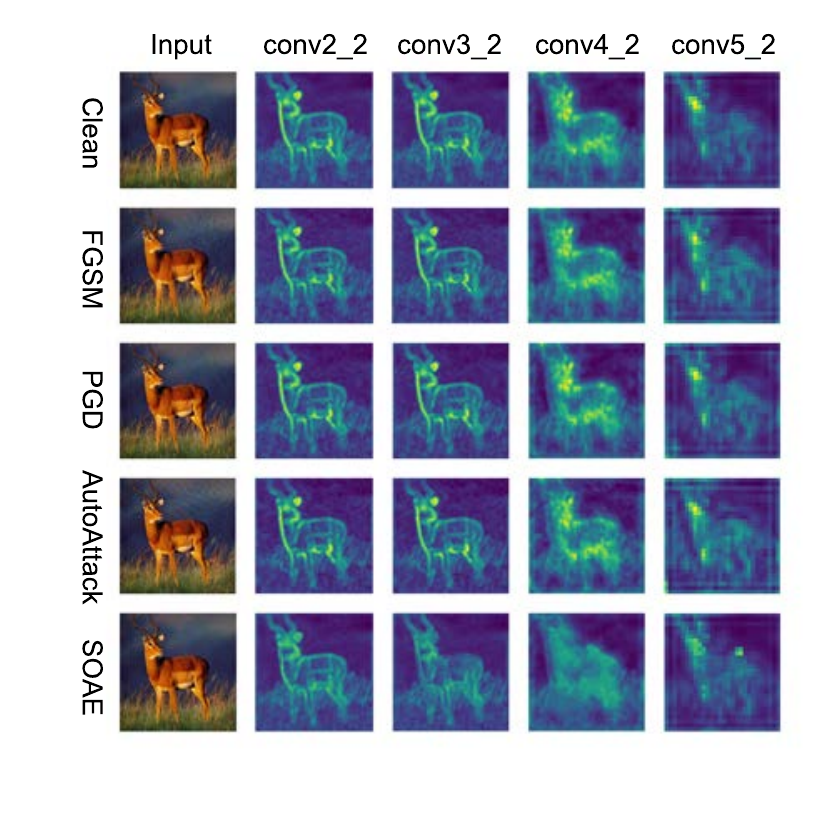}
		\caption{Visualization of the feature map in different layers of a standard-trained ResNet18 with different input examples. The perturbation size of all attack methods is $8/255$.}
		\label{fig:vis}
	\end{center}
\end{figure}

In this section, we visualize the feature maps of a standard-trained ResNet18 model with different input examples (clean, FGSM, PGD, AutoAttack and SOAE) on ImageNet dataset, as shown in Figure \ref{fig:vis}. Compared with the first-order methods, the adversarial perturbation of SOAE is more imperceptible to human eyes. Our SOAEs can deceive the model at a relatively low feature distortion level. Besides, we notice that most of the second-order perturbations occur on the objective edge where the pixel value has a more rapid change than other region of the image. This phenomenon aligns well with the role of second-order differential operator in the field of image processing, which has a stronger ability to locate the edge information. This visualization further implies the hidden relationship between the adversarial robustness and the edge information of the classification objective. According to the work of \cite{deepai}, a disruption of low-level semantic, such as edge information, can significantly weaken a CNN's ability to understand the high-level semantic. This provide a reasonable explanation for our method's effectiveness.

\begin{table*}[t]
	\caption{PSNR and SSIM of different adversarial examples.}
	\label{tab:psnr}
	\begin{center}
		\begin{small}
			\begin{tabular}{c|ccc|ccc|ccc}
				\toprule
				Method &Dataset  & PSNR  & SSIM &  Dataset & PSNR  & SSIM&  Dataset & PSNR  & SSIM\\
				\midrule
				FGSM & \multirow{4}{*}{MNIST}  & 74.23  & 0.86 
				& \multirow{4}{*}{CIFAR-10}  & 74.70  & 0.86  & 
				\multirow{4}{*}{ImageNet-100}  & 75.89& 0.87\\
				PGD &~  & 72.30  & 0.83 & ~    & 74.11 &0.84 & ~    & 78.41& 0.86\\
				AutoAttack&~ & 71.49  & 0.82&	~    &72.62 & 0.85&	~    & 77.07& 0.84\\
				SOAE &~  & \textbf{76.08}  & \textbf{0.87}&~  & \textbf{79.53}& \textbf{0.86}&~  & \textbf{81.66}& \textbf{0.89}\\
				\bottomrule
			\end{tabular}
		\end{small}
	\end{center}
\end{table*}

To further demonstrate our second-order perturbations' imperceptibility, we use Peak Signal to Noise Ratio (PSNR) \cite{PSNR} and Structural Similarity (SSIM) \cite{SSIM} to quantitatively measure the similarity and distortion between original examples and their corresponding adversarial examples. In the field of image signal processing, these two matrices are often used to evaluate the quality of the generated images, where PSNR focus more on the pixel values while SSIM comprehensively measures the structural similarity including brightness, contrast, and structure. Generally, the higher the PSNR and SSIM values are, the less distortion exists between a perturbated image and its original image. We calculate the average PSNR and SSIM of 1,000 adversarial examples (FGSM, PGD, AutoAttack, and SOAE) with respect to their original examples on MNIST, CIFAR-10 and ImageNet-100 respectively. The experiment results are reported in Table \ref{tab:psnr}. Among all adversarial attack methods, SOAEs achieve the highest PSNR and SSIM values, which indicates that our method can generate a low-distortion and imperceptible version of adversarial examples. We speculate that this attributes to the more smooth approximation of the loss function of the model, which can effectively find an adversary point within the $B_p(\epsilon)$ of the input image.

\section{Conclusion}                                                             

In this paper, we propose a Hessian-free second-order adversarial example generation method that can effectively deceive most of the first-order robust models. We further apply them to adversarial training, which effectively enable the model obtain stronger robustness against various attacks including first-order and second-order attacks. Benefit from constructing Hessian-vector product in the Krylov subspace, our algorithm avoid directly computing Hessian matrix for gradient update, which make second-order methods practical in adversarial learning. In the future, a feasible direction of our work is to better understand the landscape of the loss function around the input images, hence to select a more reasonable initial point of the Taylor expansion and make the approximation more accurate.

\bibliographystyle{IEEEtran}
\bibliography{IEEEref.bib}

\end{document}